\documentclass[11pt,a4paper]{article}
\usepackage[hyperref]{acl2020}
\usepackage{times}
\usepackage{latexsym}

\usepackage{url}
\usepackage{multirow}
\usepackage{balance}
\usepackage{subfigure}
\usepackage{latexsym}
\usepackage{amsmath}
\usepackage{amssymb}
\usepackage{caption}
\usepackage{graphicx}
\usepackage{url}
\usepackage{multicol}
\usepackage{multirow}
\usepackage{booktabs}
\usepackage{color}
\usepackage{subfigure}
\usepackage{array}
\usepackage{balance}

\usepackage{microtype}
\definecolor{midnightgreen}{rgb}{0.0, 0.29, 0.33}

\aclfinalcopy 

\title{Fine-grained Fact Verification with Kernel Graph Attention Network}

\author{Zhenghao Liu$^1$ \qquad Chenyan Xiong$^2$ \qquad Maosong Sun$^1$ \qquad Zhiyuan Liu$^1$ \\ 
$^1$Department of Computer Science and Technology, Tsinghua University, Beijing, China\\
Institute for Artificial Intelligence, Tsinghua University, Beijing, China\\
State Key Lab on Intelligent Technology and Systems, Tsinghua University, Beijing, China\\
$^2$Microsoft Research AI,  Redmond, USA\\}

\date{}

\begin{document}
\maketitle
\begin{abstract}
Fact Verification requires fine-grained natural language inference capability that finds subtle clues to identify the syntactical and semantically correct but not well-supported claims. This paper presents Kernel Graph Attention Network (KGAT), which conducts more fine-grained fact verification with kernel-based attentions. Given a claim and a set of potential evidence sentences that form an evidence graph, KGAT introduces node kernels, which better measure the importance of the evidence node, and edge kernels, which conduct fine-grained evidence propagation in the graph, into Graph Attention Networks for more accurate fact verification. KGAT achieves a 70.38\% FEVER score and significantly outperforms existing fact verification models on FEVER, a large-scale benchmark for fact verification. Our analyses illustrate that, compared to dot-product attentions, the kernel-based attention concentrates more on relevant evidence sentences and meaningful clues in the evidence graph, which is the main source of KGAT's effectiveness. All source codes of this work are available at \url{https://github.com/thunlp/KernelGAT}.
\end{abstract}
\section{Introduction}
Online contents with false information, such as fake news, political deception, and online rumors, have been growing significantly and spread widely over the past several years.
How to automatically ``fact check'' the integrity of textual contents, to prevent the spread of fake news, and to avoid the undesired social influences of maliciously fabricated statements, is urgently needed for our society.

\begin{figure}[t]
    \centering
    \includegraphics[width=0.95\linewidth]{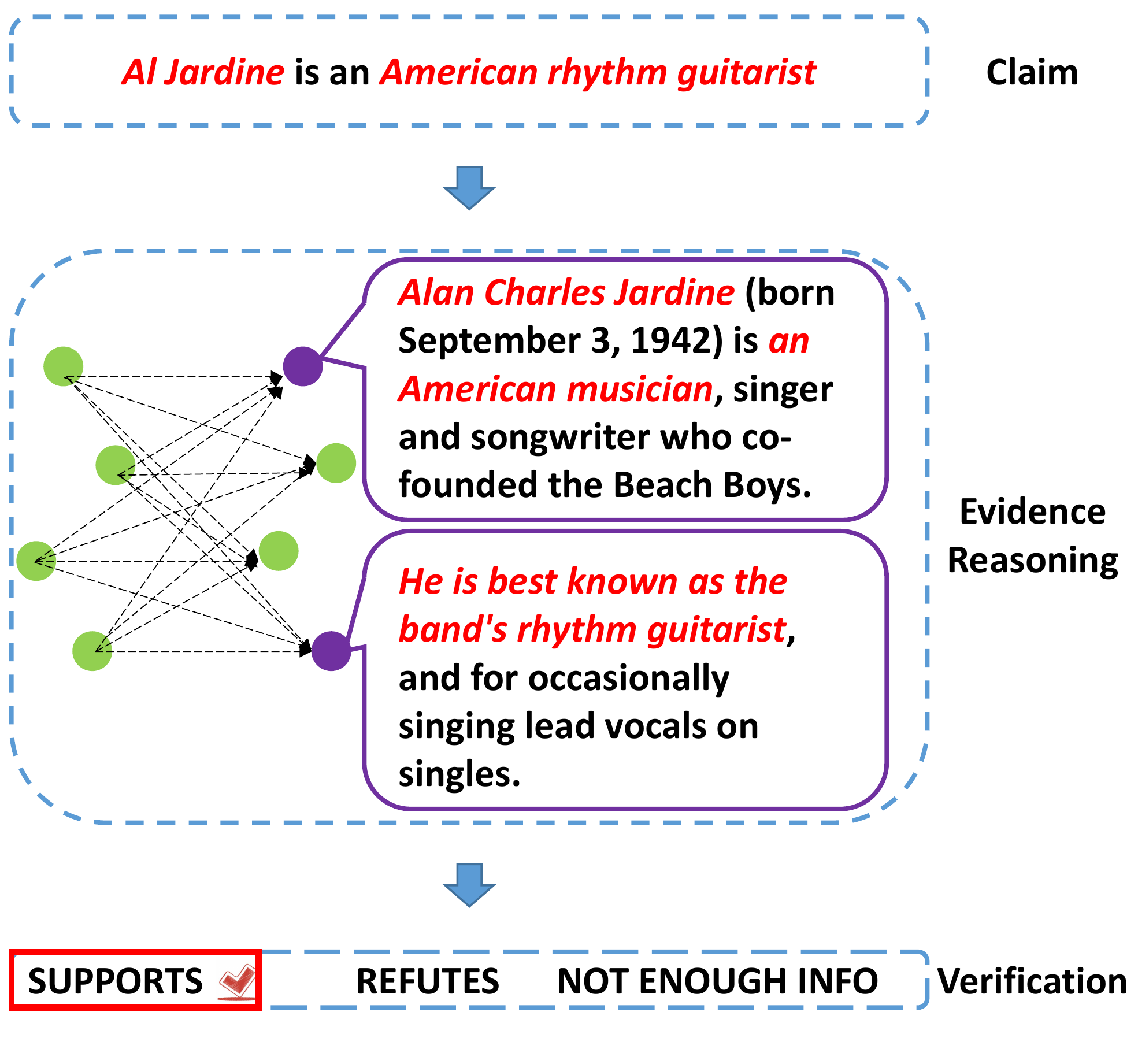}
    \caption{An Example of Fact Verification System.}
    \label{fig:example}
\end{figure}

Recent research formulates this problem as the fact verification task, which targets to automatically verify the integrity of statements using trustworthy corpora, e.g., Wikipedia~\cite{thorne2018fever}.
For example, as shown in Figure~\ref{fig:example}, a system could first retrieve related evidence sentences from the background corpus, conduct joint reasoning over these sentences, and aggregate the signals to verify the claim integrity~\cite{nie2019combining,zhou2019gear,yoneda2018ucl,hanselowski2018ukp}.

There are two challenges for evidence reasoning and aggregation in fact verification.
One is that no ground truth evidence is given; the evidence sentences are retrieved from background corpora, which inevitably contain noise.
The other is that the false claims are often deliberately fabricated; they may be semantically correct but are not supported.
This makes fact verification a rather challenging task, as it requires the fine-grained reasoning ability to distinguish the subtle differences between truth and false statements~\cite{zhou2019gear}.

This paper presents a new neural structural reasoning model, Kernel Graph Attention Network (KGAT), that provides more fine-grained evidence selection and reasoning capability for fact verification using neural matching kernels~\cite{xiong2017knrm, convknrm}.
Given retrieved evidence pieces, KGAT first constructs an evidence graph, using claim and evidence as graph nodes and fully-connected edges. It then utilizes two sets of kernels, one on the edges, which selectively summarize clues for a more fine-grained node representation and propagate clues among neighbor nodes through a multi-layer graph attention; and the other on the nodes, which performs more accurate evidence selection by better matching evidence with the claim. These signals are combined by KGAT, to jointly learn and reason on the evidence graph for more accurate fact verification.

In our experiments on FEVER~\cite{thorne2018fever}, a large-scale fact verification benchmark, KGAT achieves a 70.38\% FEVER score, significantly outperforming previous BERT and Graph Neural Network (GNN) based approaches~\cite{zhou2019gear}.
Our experiments demonstrate KGAT's strong effectiveness especially on facts that require multiple evidence reasoning: our kernel-based attentions provide more sparse and focused attention patterns, which are the main source of KGAT's effectiveness.
\section{Related Work}
The FEVER shared task~\cite{thorne2018fever} aims to develop automatic fact verification systems to check the veracity of human-generated claims by extracting evidence from Wikipedia. The recently launched FEVER shared task 1.0 is hosted as a competition on Codalab\footnote{\url{https://competitions.codalab.org/competitions/18814}} with a blind test set and has drawn lots of attention from NLP community.

Existing fact verification models usually employ FEVER's official baseline~\cite{thorne2018fever} with a three-step pipeline system~\cite{chen2017reading}: document retrieval, sentence retrieval and claim verification. Many of them mainly focus on the claim verification step.
\citet{nie2019combining} concatenates all evidence together to verify the claim. 
One can also conduct reasoning for each claim evidence pair and aggregate them to the claim label~\cite{luken2018qed,yoneda2018ucl,hanselowski2018ukp}. TwoWingOS~\cite{yin2018twowingos} further incorporates evidence identification to improve claim verification.

GEAR~\cite{zhou2019gear} formulates claim verification as a graph reasoning task and provides two kinds of attentions. It conducts reasoning and aggregation over claim evidence pairs with a graph model~\cite{velivckovic2017graph, scarselli2008graph, kipf2016semi}. \citet{zhong2019reasoning} further employs XLNet~\cite{yang2019xlnet} and establishes a semantic-level graph for reasoning for a better performance. These graph based models establish node interactions for joint reasoning over several evidence pieces.

Many fact verification systems leverage Natural Language Inference (NLI) techniques~\cite{chen2017enhanced,ghaeini2018dr,parikh2016decomposable,radford2018improving,peters2018deep,li2019several} to verify the claim. The NLI task aims to classify the relationship between a pair of premise and hypothesis as either entailment, contradiction or neutral, similar to the FEVER task, though the later requires systems to find the evidence pieces themselves and there are often multiple evidence pieces.
One of the most widely used NLI models in FEVER is Enhanced Sequential Inference Model (ESIM)~\cite{chen2017enhanced}, which employs some forms of hard or soft alignment to associate the relevant sub-components between premise and hypothesis. 
BERT, the pre-trained deep bidirectional Transformer, has also been used for better text representation in FEVER and achieved better performance~\cite{devlin2019bert,li2019several,zhou2019gear,soleimani2019bert}.

The recent development of neural information retrieval models, especially the interaction based ones, have shown promising effectiveness in extracting soft match patterns from query-document interactions~\cite{arcii, Pang2016TextMA, jiafeng2016deep, xiong2017knrm, convknrm}.
One of the effective ways to model text matches is to leverage matching kernels~\cite{xiong2017knrm, convknrm}, which summarize word or phrase interactions in the learned embedding space between query and documents. The kernel extracts matching patterns which provide a variety of relevance match signals and shows strong performance in various ad-hoc retrieval dataset~\cite{dai2019deeper}. Recent research also has shown kernels can be integrated with contextualized representations, i.e., BERT, to better model the relevance between query and documents~\cite{macavaney2019cedr}.
\section{Kernel Graph Attention Network}
\begin{figure}[t]
    \centering
    \includegraphics[width=0.95\linewidth]{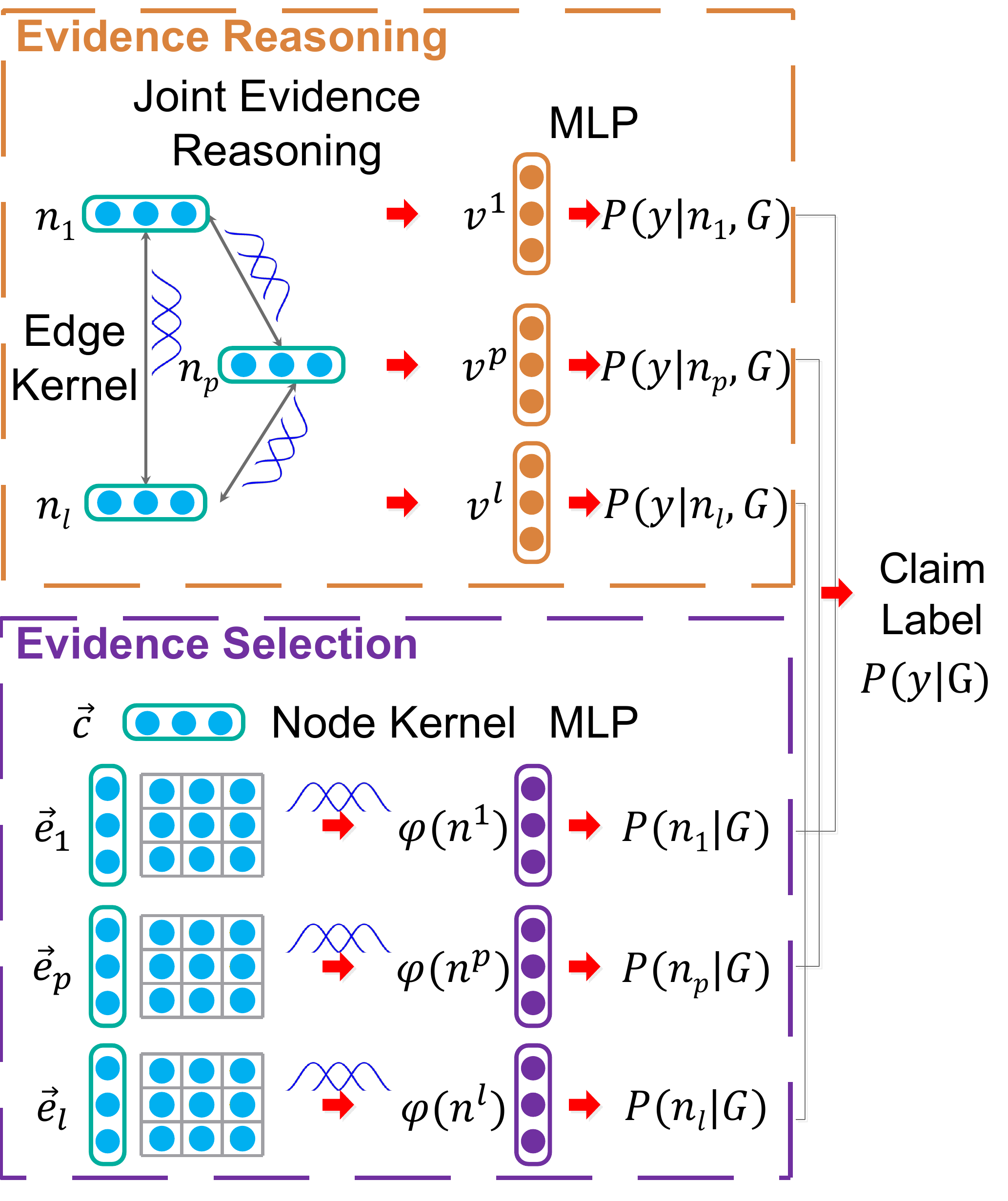}
    \caption{KGAT Architecture.}
    \label{fig:model}
\end{figure}

This section describes our Kernel Graph Attention Network (KGAT) and its application in Fact Verification. 
Following previous research, KGAT first constructs an evidence graph using retrieved evidence sentences $D=\{e^1, \dots, e^p , \dots, e^l \}$ for claim $c$, and then uses the evidence graph to predict the claim label $y$ (Sec.~\ref{sec:graph} and~\ref{model:initial}). As shown in Figure~\ref{fig:model}, the reasoning model includes two main components: Evidence Propagation with Edge Kernels (Sec.~\ref{model:edge_kernel}) and Evidence Selection with Node Kernels (Sec.~\ref{model:node_kernel}).

\subsection{Reasoning with Evidence Graph}\label{sec:graph}
Similar to previous research~\citep{zhou2019gear},
KGAT constructs the evidence graph $G$ by using each claim-evidence pair as a node and connects all node pairs with edges, making it a fully-connected evidence graph with $l$ nodes: $N=\{n^1,\dots,n^p, \dots, n^l\}$.

KGAT unifies both multiple and single evidence reasoning scenarios and produces a probability $P(y|c, D)$ to predict claim label $y$. Different from previous work~\cite{zhou2019gear}, we follow the standard graph label prediction setting in graph neural network~\cite{velivckovic2017graph} and split the prediction into two components: 1) the label prediction in each node conditioned on the whole graph $ P(y|n^p, G)$; 2) the evidence selection probability $P(n^p|G)$:
\begin{equation}
\small
P(y|c, D) = \sum_{p=1}^{l} P(y|c, e^p, D) P(e^p|c, D),
\end{equation}
or in the graph notation:
\begin{equation}
\small
P(y|G) = \sum_{p=1}^{l} P(y|n^p, G) P(n^p|G).
\end{equation}
The joint reasoning probability $P(y|n^p, G)$ calculates node label prediction with multiple evidence. 
The readout module~\cite{knyazev2019understanding} calculates the probability $P(n^p|G)$ and attentively combines per-node signals for prediction. 

The rest of this section describes the initialization of node representations ($n^p$) in Sec.~\ref{model:initial}, the calculation of per-node predictions $P(y|n^p, G)$ with Edge Kernels (Sec.~\ref{model:edge_kernel}), and the readout module $P(n^p|G)$ with Node Kernels (Sec.~\ref{model:node_kernel}).

\subsection{Initial Node Representations}\label{model:initial}
The node representations are initialized by feeding the concatenated sequence of claim, document (Wiki) title, and evidence sentence, to pre-trained BERT model~\cite{devlin2019bert}. 
Specifically, in the node $n_p$, the claim and evidence correspond to $m$ tokens (with ``[SEP]'') and $n$ tokens (with Wikipedia title and ``[SEP]'') . Using the BERT encoder, we get the token hidden states $H^{p}$ with the given node $n^p$:
\begin{equation}
\small
H^{p} = \text{BERT}(n^p).
\end{equation}

The representation of the first token (``[CLS]'') is denoted as the initial representation of node $n^p$:
\begin{equation}
\small
z^{p} = H^{p}_0.
\end{equation}

The rest of the sequences $H^{p}_{1:m+n}$ are also used to represent the claim and evidence tokens: $H^{p}_{1:m}$ for the claim tokens and  $H^{p}_{m+1:m+n}$ for the evidence tokens.

\subsection{Edge Kernel for Evidence Propagation}\label{model:edge_kernel}
The evidence propagation and per-node label prediction in KGAT are conducted by Edge Kernels, which attentively propagate information among nodes in the graph $G$ along the edges with the kernel attention mechanism. 

Specifically, KGAT calculates the node $n^p$'s representation $v^p$ with the kernel attention mechanism, and uses it to produce the per-node claim prediction $y$:
\begin{equation}
\small
\begin{aligned}
v^p &= \text{Edge-Kernel} (n^p, G), \\
P(y|n^p, G) &= \text{softmax}_y (\text{Linear}( v^p)).
\end{aligned}
\end{equation}

The edge kernel of KGAT conducts a hierarchical attention mechanism to propagate information between nodes. It uses \textit{token level attentions} to produce node representations and \textit{sentence level attentions} to propagate information along edges.

\textbf{Token Level Attention.} The token level attention uses kernels to get the fine-grained representation $\hat{z}^{q \rightarrow p}$ of neighbor node $n^q$, according to node $n^p$. The content propagation and the attention are controlled by kernels.

To get the attention weight $\alpha_i^{q \rightarrow p}$ for $i$-th token in $n^q$,
we first conduct a translation matrix $M^{q \rightarrow p}$ between $q$-th node and $p$-th node. Each element of the translation matrix $M_{ij}^{q \rightarrow p}$ in $M^{q \rightarrow p}$ is the cosine similarity of their corresponding tokens' BERT representations:
\begin{equation}
\small
M_{ij}^{q \rightarrow p} = \cos (H^q_i, H^p_j).
\end{equation}
Then we use $K$ kernels to extract the matching feature $\vec{K}(M^{q \rightarrow p}_i)$ from the translation matrix $M^{q \rightarrow p}$~\citep{xiong2017knrm,convknrm,qiao2019understanding,macavaney2019cedr}:
\begin{equation}
\small
\vec{K}(M^{q \rightarrow p}_i) = \{ K_1(M^{q \rightarrow p}_i), ... ,K_K(M^{q \rightarrow p}_i) \}.
\end{equation}
Each kernel $K_k$ utilizes a Gaussian kernel to extract features and summarizes the translation score to support multi-level interactions: 
\begin{equation}
\small
K_k(M^{q \rightarrow p}_i) = \log \sum_j \exp (- \frac{(M^{q \rightarrow p}_{ij}-\mu_k)^2}{2 \delta_k^2}),
\end{equation}
where $\mu_k$ and $\delta_k$ are the mean and width for the $k$-th kernel, which captures a certain level of interactions between the tokens~\citep{xiong2017knrm}. 

Then each token's attention weight $\alpha_i^{q \rightarrow p}$ is calculated using a linear layer:
\begin{equation}
\small
\alpha_i^{q \rightarrow p} = \text{softmax}_i (\text{Linear}( \vec{K} (M_i^{q \rightarrow p}))). \label{eq.tokenatt}
\end{equation}

The attention weights are used to combine the token representations ($\hat{z}^{q \rightarrow p}$):
\begin{equation}
\small
\hat{z}^{q \rightarrow p} = \sum_{i=1}^{m+n} \alpha_i^{q \rightarrow p} \cdot H_i^{q},
\end{equation}
which encodes the content signals to propagate from node $n^q$ to node $n^p$.

\textbf{Sentence Level Attention.}
The sentence level attention combines neighbor node information to node representation $v^p$.
The aggregation is done by a graph attention mechanism, the same with previous work~\citep{zhou2019gear}.

It first calculate the attention weight $\beta^{q \rightarrow p}$ of $n^q$ node according to the $p$-th node $n^p$:
\begin{equation}
\small
\beta^{q \rightarrow p} = \text{softmax}_q (\text{MLP} (z^{p} \circ \hat{z}^{q \rightarrow p})), \label{eq.nodeatt}
\end{equation} 
where $\circ$ denotes the concatenate operator and $z^{p}$ is the initial representation of $n^p$.

Then the $p$-th node's representation is updated by combining the neighbor node representations $\hat{z}^{q \rightarrow p}$ with the attention:
\begin{equation}
\small
v^p = (\sum_{q = 1}^{l} \beta^{q \rightarrow p} \cdot \hat{z}^{q \rightarrow p})\circ z^{p}.
\end{equation}
It updates the node representation with its neighbors, and the updated information are selected first by the token level attention (Eq.~\ref{eq.tokenatt}) and then the sentence level attention (Eq.~\ref{eq.nodeatt}).

\textbf{Sentence Level Claim Label Prediction.} The updated $p$-th node representation $v^p$ is used to calculate the claim label probability $P(y|n^p)$:
\begin{equation}
\small
P(y|n^p, G) = \text{softmax}_y (\text{Linear}( v^p)).
\end{equation}
The prediction of the label probability for each node is also conditioned on the entire graph $G$, as the node representation is updated by gather information from its graph neighbors.

\subsection{Node Kernel for Evidence Aggregation}\label{model:node_kernel}
The per-node predictions are combined by the ``readout'' function in graph neural networks~\citep{zhou2019gear}, where KGAT uses node kernels to learn the importance of each evidence.

It first uses node kernels to calculate the readout representation $\phi(n^p)$ for each node $n^p$:
\begin{equation}
\small
\phi(n^p) = \text{Node-Kernel} (n^p).
\end{equation}

Similar to the edge kernels, we first conduct a translation matrix $M^{c \rightarrow e^p}$ between the $p$-th claim and evidence, using their hidden state set $H^p_{1:m}$ and  $ H^p_{m+1:m+n}$. The kernel match features  $\vec{K} (M^{c \rightarrow e^p}_i)$ on the translation matrix are combined to produce the node selection representation $\phi(n^p)$:
\begin{equation}
\small
\phi(n^p) = \frac{1}{m} \cdot \sum_{i=1}^m \vec{K} (M^{c \rightarrow e^p}_i).
\end{equation}

This representation is used in the readout to calculate $p$-th evidence selection probability $P(n^p|G)$:
\begin{equation}
\small
P(n^p|G) = \text{softmax}_p (\text{Linear} (\phi(n^p))).
\end{equation}

KGAT leverages the kernels multi-level soft matching capability~\citep{xiong2017knrm} to weight the node-level predictions in the evidence graph based on their relevance with the claim:
\begin{equation}
\small
P(y|G) = \sum_{p=1}^{l} P(y|n^p, G) P(n^p|G).
\end{equation}
The whole model is trained end-to-end by minimizing the cross entropy loss:
\begin{equation}
\small
L = \text{CrossEntropy}(y^*, P(y|G)),
\end{equation}
using the ground truth verification label $y^*$.

\section{Experimental Methodology}
This section describes the dataset, evaluation metrics, baselines, and implementation details in our experiments.

\textbf{Dataset.}
A large scale public fact verification dataset FEVER~\cite{thorne2018fever} is used in our experiments. The FEVER consists of 185,455 annotated claims with 5,416,537 Wikipedia documents from the June 2017 Wikipedia dump. All claims are classified as SUPPORTS, REFUTES or NOT ENOUGH INFO by annotators. The dataset partition is kept the same with the FEVER Shared Task~\cite{thorne2018fact} as shown in Table~\ref{tab:dataset}.

\textbf{Evaluation Metrics.}
The official evaluation metrics\footnote{\url{https://github.com/sheffieldnlp/fever-scorer}} for claim verification include Label Accuracy (LA) and FEVER score. LA is a general evaluation metric, which calculates claim classification accuracy rate without considering retrieved evidence. The FEVER score considers whether one complete set of golden evidence is provided and better reflects the inference ability.

We also evaluate Golden FEVER (GFEVER) scores, which is the FEVER score but with golden evidence provided to the system, an easier setting. Precision, Recall and F1 are used to evaluate evidence sentence retrieval accuracy using the provided sentence level labels (whether the sentence is evidence or not to verify the claim).

\textbf{Baselines.}
The baselines include top models during FEVER 1.0 task and BERT based models.

Three top models in FEVER 1.0 shared task are compared. Athene~\cite{hanselowski2018ukp} and UNC NLP~\cite{nie2019combining} utilize ESIM to encode claim evidence pairs. UCL MRG~\cite{yoneda2018ucl} leverages Convolutional Neural Network (CNN) to encode claim and evidence. These three models aggregate evidence by attention mechanism or label aggregation component.

The BERT based models are our main baselines, they significantly outperform previous methods without pre-training. BERT-pair, BERT-concat and GEAR are three baselines from the previous work~\cite{zhou2019gear}. BERT-pair and BERT-concat regard claim-evidence pair individually or concatenate all evidence together to predict claim label. GEAR utilizes a graph attention network to extract supplement information from other evidence and aggregate all evidence through an attention layer.
\citet{soleimani2019bert, nie2019revealing} are also compared in our experiments. They implement BERT sentence retrieval for a better performance.  
In addition, we replace kernel with dot product to implement our GAT version, which is similar to GEAR, to evaluate kernel's effectiveness.

\begin{table}[t]
\begin{center}
\resizebox{0.49\textwidth}{!}{
\begin{tabular}{c  c  c  c}
\hline \textbf{Split} & \textbf{SUPPORTED} & \textbf{REFUTED} & \textbf{NOT ENOUGH INFO}\\ \hline
\textbf{Train} & 80,035 & 29,775 & 35,639  \\
\textbf{Dev} & 6,666 & 6,666 & 6,666  \\
\textbf{Test} & 6,666 & 6,666 & 6,666  \\ \hline
\end{tabular}}
\caption{\label{tab:dataset}Statistics of FEVER Dataset.}
\end{center}
\end{table}

\textbf{Implementation Details.} The rest of this section describes our implementation details. 

\textit{Document retrieval.} The document retrieval step retrieves related Wikipedia pages and is kept the same with previous work~\cite{hanselowski2018ukp,zhou2019gear,soleimani2019bert}. For a given claim, it first utilizes the constituency parser in AllenNLP~\cite{gardner2018allennlp} to extract all phrases which potentially indicate entities. Then it uses these phrases as queries to find relevant Wikipedia pages through the online MediaWiki API\footnote{\url{https://www.mediawiki.org/wiki/API: Main_page}}. Then the convinced article are reserved~\cite{hanselowski2018ukp}.

\textit{Sentence retrieval.} The sentence retrieval part focuses on selecting related sentences from retrieved pages. There are two sentence retrieval models in our experiments: ESIM based sentence retrieval and BERT based sentence retrieval. The ESIM based sentence retrieval keeps the same as the previous work~\cite{hanselowski2018ukp,zhou2019gear}.
The base version of BERT is used to implement our BERT based sentence retrieval model. We use the ``[CLS]'' hidden state to represent claim and evidence sentence pair. Then a learning to rank layer is leveraged to project ``[CLS]'' hidden state to ranking score. Pairwise loss is used to optimize the ranking model. Some work~\cite{Zhao2020transformerxh,Ye2020coreferentialRL} also employs our BERT based sentence retrieval in their experiments.

\textit{Claim verification.} During training, we set the batch size to 4 and accumulate step to 8. All models are evaluated with LA on the development set and trained for two epochs. The training and development sets are built with golden evidence and higher ranked evidence with sentence retrieval. All claims are assigned with five pieces of evidence.
The BERT (Base), BERT (Large) and RoBERTa~\cite{liu2019roberta} are evaluated in claim verification.

In our experiments, the max length is set to 130. All models are implemented with PyTorch. BERT inherits huggingface's implementation\footnote{\url{https://github.com/huggingface/pytorch-transformers}}. Adam optimizer is used with learning rate = 5e-5 and warm up proportion = 0.1. The kernel size is set to 21, the same as previous work~\cite{qiao2019understanding}.

\section{Evaluation Result}
The experiments are conducted to study the performance of KGAT, its advantages on different reasoning scenarios, and the effectiveness of kernels.

\begin{table}[t]
\begin{center}
\resizebox{0.49\textwidth}{!}{
\begin{tabular}{l  c  c | c  c}
\hline \multirow{2}{*}{\textbf{Model}} & \multicolumn{2}{c|}{\textbf{Dev}} &  \multicolumn{2}{c}{\textbf{Test}} \\ \cline{2-5}
 & \textbf{LA} & \textbf{FEVER} & \textbf{LA} & \textbf{FEVER} \\ \hline
Athene~\cite{hanselowski2018ukp} & 68.49 & 64.74 & 65.46 & 61.58 \\ 
UCL MRG~\cite{yoneda2018ucl} &  69.66 & 65.41 & 67.62 & 62.52\\ 
UNC NLP~\cite{nie2019combining} &  69.72 & 66.49 & 68.21 & 64.21 \\ \hline
BERT Concat~\cite{zhou2019gear} & 73.67 & 68.89 & 71.01 & 65.64 \\
BERT Pair~\cite{zhou2019gear} & 73.30 & 68.90 & 69.75 & 65.18\\
GEAR~\cite{zhou2019gear} & 74.84 & 70.69 & 71.60 & 67.10 \\
GAT (BERT Base) w. ESIM Retrieval & 75.13 & 71.04 & 72.03 & 67.56 \\
KGAT (BERT Base) w. ESIM Retrieval& \textbf{75.51} & \textbf{71.61} & \textbf{72.48} & \textbf{68.16} \\\hline
SR-MRS~\cite{nie2019revealing} & 75.12 & 70.18 & 72.56 & 67.26\\
BERT (Base)~\cite{soleimani2019bert} & 73.51 & 71.38 & 70.67 & 68.50\\
KGAT (BERT Base) & \textbf{78.02} & \textbf{75.88} & \textbf{72.81} & \textbf{69.40} \\ \hline
BERT (Large)~\cite{soleimani2019bert} & 74.59 & 72.42  & 71.86 & 69.66\\
KGAT (BERT Large) & \textbf{77.91} & \textbf{75.86} & \textbf{73.61}  & \textbf{70.24} \\ \hline
KGAT (RoBERTa Large) & \textbf{78.29} & \textbf{76.11} & \textbf{74.07}  & \textbf{70.38} \\
\hline
\end{tabular}}
\caption{\label{tab:pipeline}Fact Verification Accuracy. The performances of top models during FEVER 1.0 shared task and BERT based models with different scenarios are presented.}
\end{center}
\end{table}

\subsection{Overall Performance}
The fact verification performances are shown in
Table~\ref{tab:pipeline}. Several testing scenarios are conducted to compare KGAT effectiveness to BERT based baselines: BERT (Base) Encoder with ESIM retrieved sentences, with BERT retrieved sentences, and BERT (Large) Encoder with BERT retrieved sentences.


Compared with baseline models, KGAT is the best on all testing scenarios.
With ESIM sentence retrieval, same as the previous work~\cite{zhou2019gear,hanselowski2018ukp}, KGAT outperforms the graph attention models GEAR and our GAT on both development and testing sets. It illustrates the effectiveness of KGAT among graph based reasoning models. With BERT based sentence retrieval, our KGAT also outperforms BERT (Base)~\cite{soleimani2019bert} by almost 1\% FEVER score, showing consistent effectiveness with different sentence retrieval models. When using BERT (Large) as the encoder, KGAT also outperforms the corresponding version of~\citet{soleimani2019bert}.
KGAT with RoBERTa performs the best compared with all previously published research on all evaluation metrics. CorefBERT~\cite{Ye2020coreferentialRL} extends our KGAT architecture and explicitly models co-referring relationship in context for better performance.

The sentence retrieval performances of ESIM and BERT are compared in Table~\ref{tab:retrieval}.
The BERT sentence retrieval outperforms ESIM sentence retrieval significantly, thus also helps improve KGAT's reasoning accuracy. Nevertheless, for more fair comparisons, our following experiments are all based on ESIM sentence retrieval, which is the one used by GEAR, our main baseline~\cite{zhou2019gear}.

\begin{table}[t]
\begin{center}
\small
\resizebox{0.45\textwidth}{!}{
\begin{tabular}{l l  c  c  c | c}
\hline  & \textbf{Model} & \textbf{Prec@5} & \textbf{Rec@5} & \textbf{F1@5}&  \textbf{FEVER}\\ \hline
\multirow{2}{*}{\textbf{Dev}} & ESIM    & 24.08 & 86.72 & 37.69 & 71.70\\ 
&BERT&  \textbf{27.29} & \textbf{94.37} & \textbf{42.34} & \textbf{75.88}\\ \hline
\multirow{2}{*}{\textbf{Test}} & ESIM    & 23.51 & 84.66 & 36.80 & 68.16\\ 
&BERT &  \textbf{25.21} & \textbf{87.47} & \textbf{39.14} & \textbf{69.40}\\
\hline
\end{tabular}}
\caption{\label{tab:retrieval}Evidence Sentence Retrieval Accuracy. Sentence level \textbf{Prec}ision, \textbf{Rec}all and \textbf{F1} are evaluated by official evaluation~\cite{thorne2018fever}.
}
\end{center}
\end{table}

\subsection{Performance on Different Scenarios}
This experiment studies the effectiveness of kernel on multiple and single evidence reasoning scenarios, as well as the contribution of kernels.

The verifiable instances are separated (except instances with ``NOT ENOUGH INFO'' label ) into two groups according to the golden evidence labels. If more than one evidence pieces are required, the claim is considered as requiring multi-evidence reasoning. The single evidence reasoning set and the multiple evidence reasoning set contain 11,372 (85.3\%) and 1,960 (14.7\%) instances, respectively. 
We also evaluate two additional KGAT variations: KGAT-Node which only uses kernels on the node, with the edge kernels replaced by standard dot-production attention, and KGAT-Edge which only uses kernels on the edge. The results of these systems on the two scenarios are shown in Table~\ref{tab:ablation}.

\begin{table}[t]
\begin{center}
\resizebox{0.49\textwidth}{!}{
\begin{tabular}{c l c c c c}
\hline \textbf{Reasoning} & \textbf{Model} & \textbf{LA} & \textbf{GFEVER} & \multicolumn{2}{c}{\textbf{FEVER}} \\ \hline
\multirow{5}{*}{\textbf{Multiple}} &GEAR &  \textbf{66.38} & n.a. & 37.96 & -0.25\%\\ 
&GAT &  66.12 & 84.39 & 38.21 & -\\ 
&KGAT-Node &  65.51 & 83.88 & 38.52 & 0.31\% \\
&KGAT-Edge &  65.87 & 84.90 & 39.08 & 0.87\% \\
&KGAT-Full &  65.92 & \textbf{85.15} & \textbf{39.23} & 1.02\%\\ \hline
\multirow{5}{*}{\textbf{Single}}&GEAR &  78.14 & n.a. & 75.73 & -1.69\%\\
&GAT &  79.79 & 81.96 & 77.42 & -\\ 
&KGAT-Node &  79.92 & 82.29 & 77.73 & 0.31\%\\
&KGAT-Edge &  79.90 & 82.41 & 77.58 & 0.16\%\\
&KGAT-Full &  \textbf{80.33} & \textbf{82.62} & \textbf{78.07} & 0.65\%\\
\hline
\end{tabular}}
\end{center}
\caption{\label{tab:ablation} Claim Verification Accuracy on Claims that requires Multiple and Single evidence Pieces. Standard GAT with no kernel (GAT), with only node kernel (KGAT-Node), with only edge kernel (KGAT-Edge) and the full model (KGAT-Full) are compared.
}
\end{table}
KGAT-Node outperforms GAT by more than 0.3\% on both single and multiple reasoning scenarios. As expected, it does not help much on GFEVER, because the golden evidence is given and node selection is not required. It illustrates KGAT-Node mainly focuses on choosing appropriate evidence and assigning accurate combining weights in the readout.

KGAT-Edge outperforms GAT by more than 0.8\% and 0.1\% on multiple and single evidence reasoning scenarios, respectively. Its effectiveness is mostly on combining the information from multiple evidence pieces.

The multiple and single evidence reasoning scenarios evaluate the reasoning ability from different aspects. The single evidence reasoning mainly focuses on selecting the most relevant evidence and inference with single evidence. It mainly evaluates model de-noising ability with the retrieved evidence. The multiple evidence reasoning is a harder and more complex scenario, requiring models to summarize necessary clues and reason over multiple evidence. It emphasizes to evaluate the evidence interactions for the joint reasoning. KGAT-Node shows consistent improvement on both two reasoning scenarios, which demonstrates the important role of evidence selection. KGAT-Edge, on the other hand, is more effective on multiple reasoning scenarios as the Edge Kernels help better propagate information along the edges.


\subsection{Effectiveness of Kernel in KGAT}
This set of experiments further illustrate the influences of kernels in KGAT.

\begin{figure}[t]
    \centering
    \subfigure[Edge Attention.] { \label{fig:kernel:att:entropy_edge} 
    \includegraphics[width=0.45\linewidth]{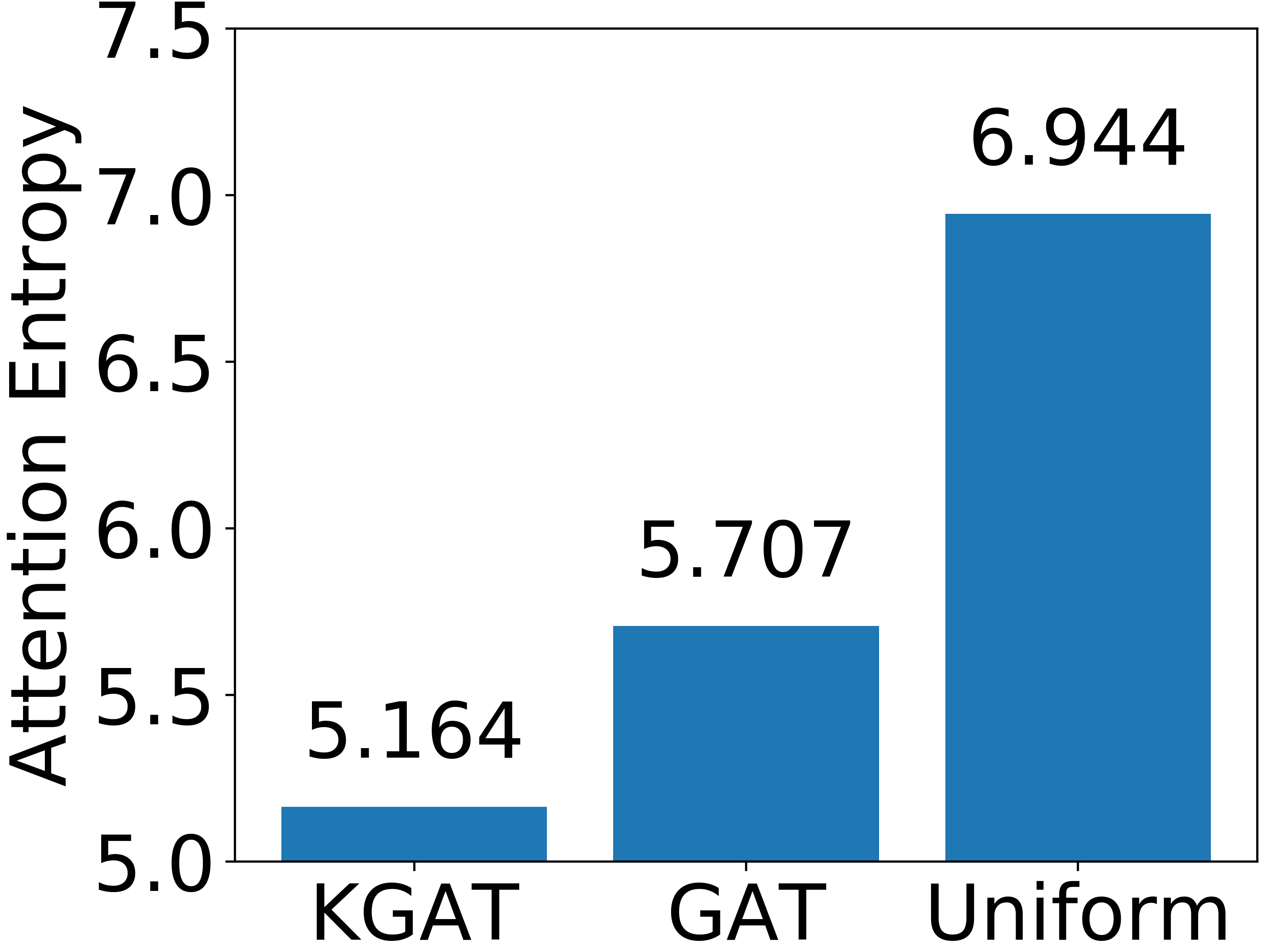}}
    \subfigure[Node Attention.] { \label{fig:kernel:att:entropy_node} 
    \includegraphics[width=0.45\linewidth]{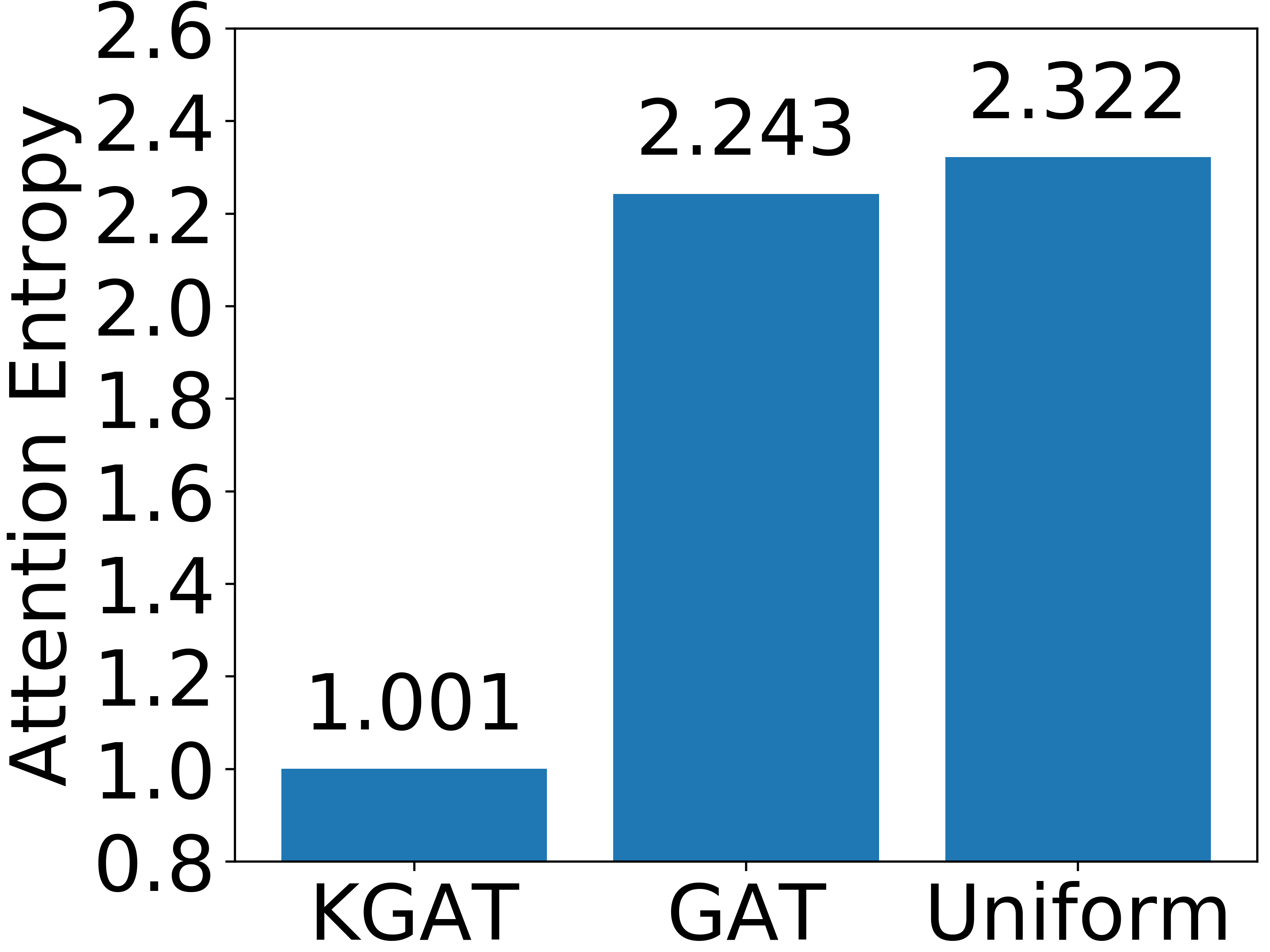}}
    \caption{Attention Weight Entropy on Evidence Graph, from KGAT and GAT, of graph edges and nodes. Uniform weights' entropy is also shown for comparison. Less entropy shows more concentrated attention.
    }
    \label{fig:kernel:att}
\end{figure}

\textbf{More Concentrated Attention.}
This experiment studies kernel attentions by their entropy, which reflects whether the learned attention weights are focused or scattered.
The entropy of the kernel attentions in KGAT, the dot-product attentions in GAT, and the uniform attentions are shown in Figure~\ref{fig:kernel:att}.

The entropy of Edge attention is shown in Figure~\ref{fig:kernel:att:entropy_edge}. Both GAT and KGAT show a smaller entropy of the token attention than the uniform distribution. It illustrates that GAT and KGAT have the ability to assign more weight to some important tokens with both dot product based and kernel based attentions. Compared to the dot-product attentions in GAT, KGAT's Edge attention focuses on fewer tokens and has a smaller entropy.

The entropy of Node attentions are plotted in Figure~\ref{fig:kernel:att:entropy_node}. GAT's attentions distribute almost the same with the uniform distribution, while KGAT has concentrated Node attentions on a few evidence sentences. As shown in the next experiment, the kernel based node attentions focus on the correct evidence pieces and de-noises the retrieved sentences, which are useful for claim verification.

\begin{figure}[t]
    \centering
    \subfigure[Attention Distribution.] { \label{fig:kernel:select:token} 
    \includegraphics[width=0.45\linewidth]{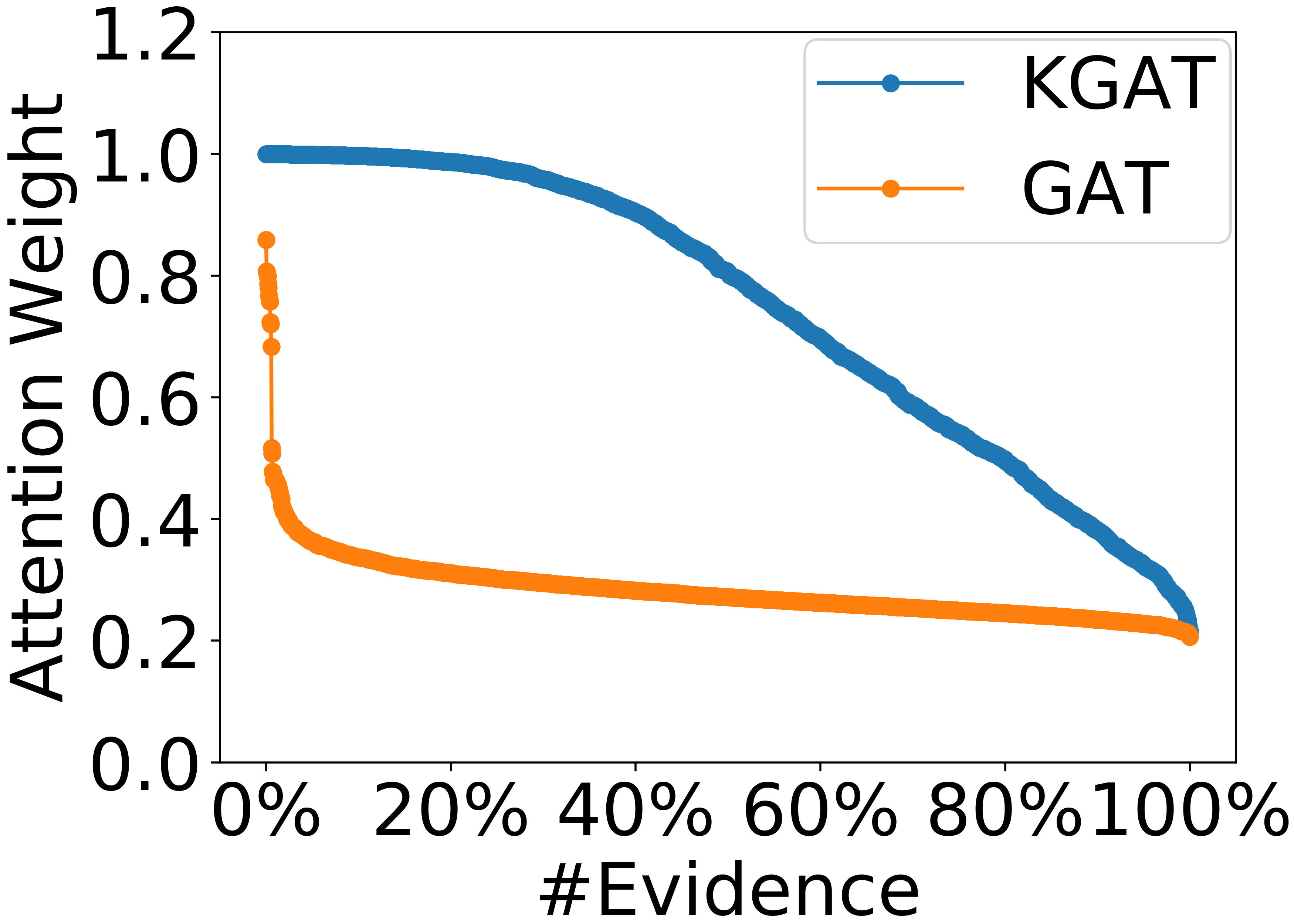}}
    \subfigure[Evidence Recall.] { \label{fig:kernel:select:recall} 
    \includegraphics[width=0.45\linewidth]{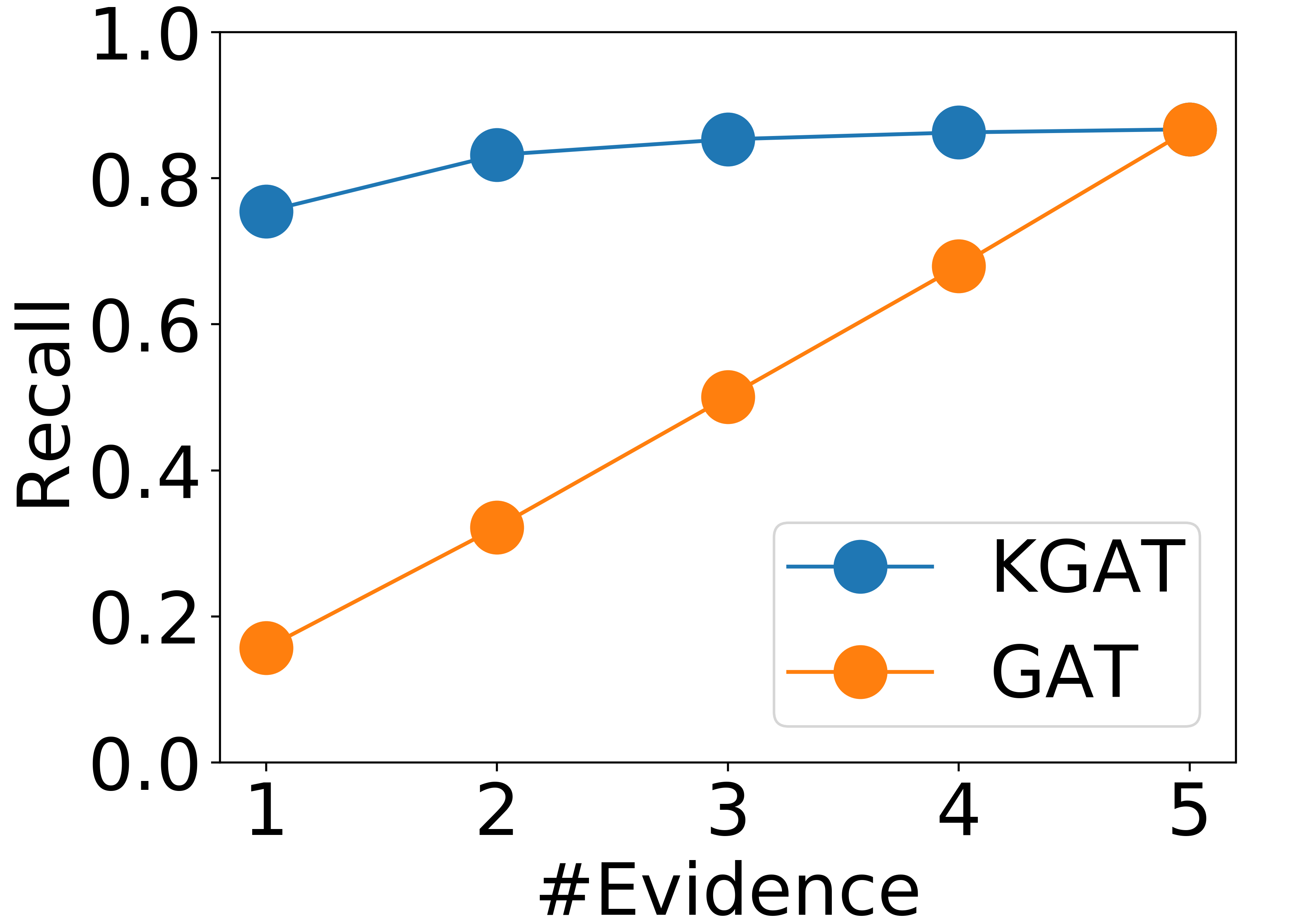}}
    \caption{Evidence Selection Effectiveness of KGAT and GAT. Fig~\ref{fig:kernel:select:token} shows the distribution of attention weights on evidence nodes $p(n^p)$, sorted by their weights; Fig~\ref{fig:kernel:select:recall} evaluates the recall of selecting the golden standard evidence nodes at different depths.}
    \label{fig:kernel:select}
\end{figure}
\begin{figure}[t]
    \centering
    \subfigure[GAT.] { \label{fig:kernel:token:dot} 
    \includegraphics[width=0.45\linewidth]{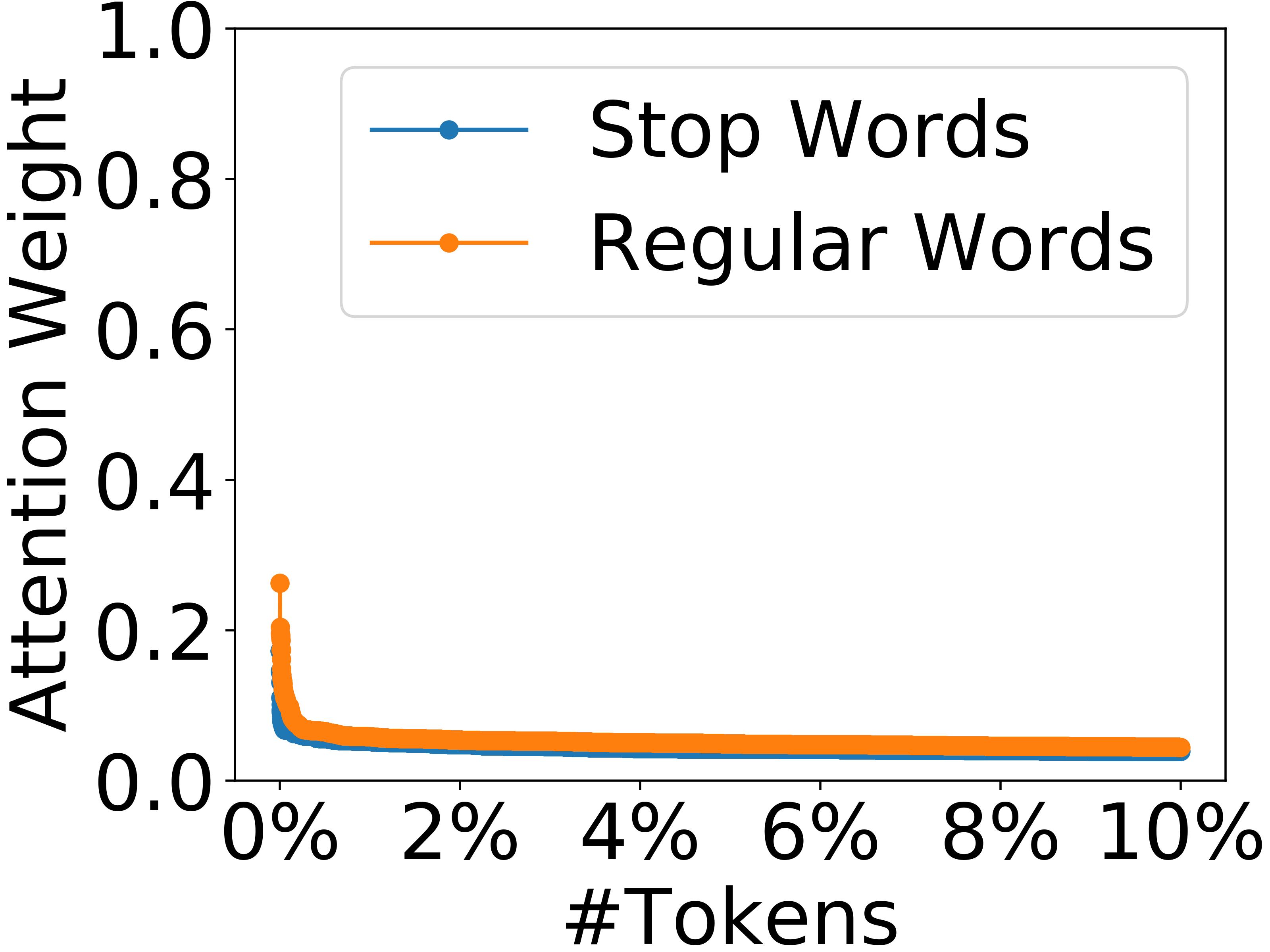}}
    \subfigure[KGAT.] { \label{fig:kernel:token:kernel} 
    \includegraphics[width=0.45\linewidth]{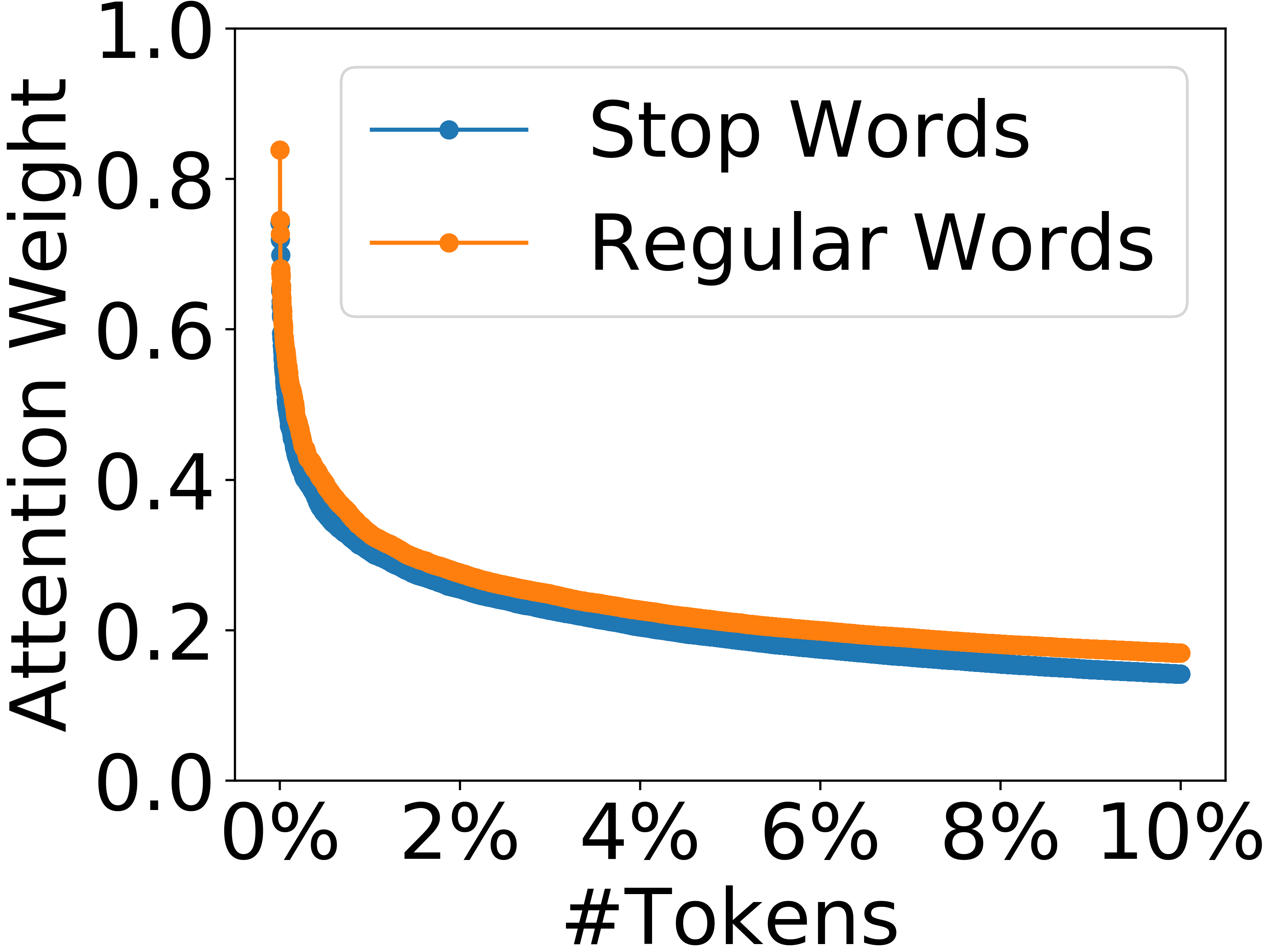}}
    \caption{The Attention Weight Distribution from GAT and KGAT on evidence sentence tokens. Top 10\% tokens are presented. The rest follows standard long tail distributions.}
    \label{fig:kernel:token}
\end{figure}


\textbf{More Accurate Evidence Selection.}
This experiment evaluates the effectiveness of KGAT-Node through attention distribution and evidence recall. The results are shown in Figure~\ref{fig:kernel:select}.

We first obtain the node attention score in the evidence graph from KGAT or GAT, and calculate the statistics of the maximum one for each claim, as most of which only require single evidence to verify. The attention score of the highest attended evidence node for each claim is plotted in Figure~\ref{fig:kernel:select:token}.
As expected,  KGAT concentrates its weight to select evidence nodes and provides a focused attention.


Then the evidence selection accuracy is evaluated by their evidence recall. We first rank all evidence pieces for each claim. Then the evidence recall with different ranking depths is plotted in Figure~\ref{fig:kernel:select:recall}. KGAT achieves a much higher recall on top ranking positions---only the first ranked sentence covers nearly 80\% of ground truth evidence, showing the node kernels' ability to select correct evidence. This also indicates the potential of the node kernels in the sentence retrieval stage, which we reserve for future work as this paper focuses on the reasoning stage.

\textbf{Fine-Grained Evidence Propagation.} 
The third analysis studies the distribution of KGAT-Edge's attention which is used to propagate the evidence clues in the evidence graph.

Figure~\ref{fig:kernel:token} plots the attention weight distribution of the edge attention scores in KGAT and GAT, one from kernels and one from dot-products.
The kernel attentions again are more concentrated: KGAT focuses fewer words while GAT's dot-product attentions are almost equally distributed among all words. This observation of the scattered dot-product attention is consistent with previous research~\cite{clark2019does}.
As shown in the next case study, the edge kernels provide a fine-grained and intuitive attention pattern when combining evidence clues from multiple pieces.

\section{Case Study}
\begin{figure}[t]
    \centering
    \includegraphics[width=0.95\linewidth]{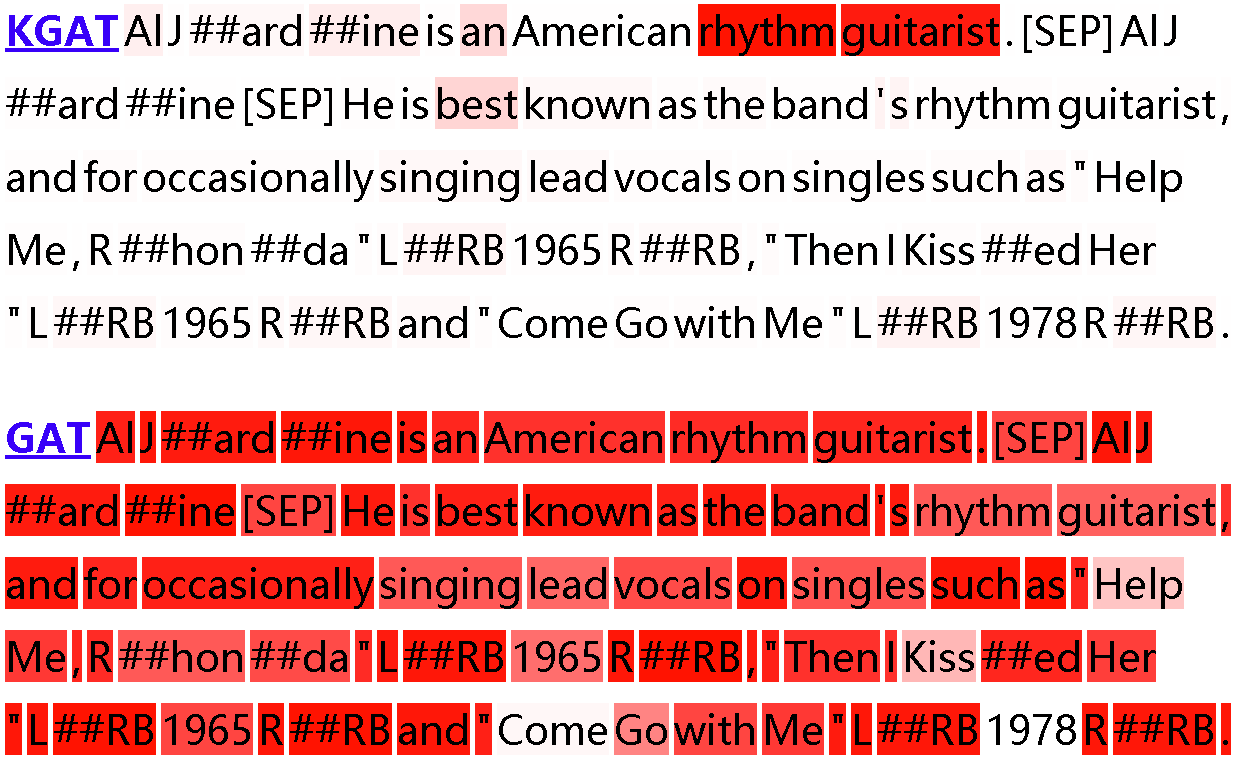}
    \caption{Edge Attention Weights on Evidence Tokens. Darker red indicates higher attention weights. 
    }
    \label{fig:case:token}
\end{figure}

Table~\ref{tab:case} shows the example claim used in GEAR~\cite{zhou2019gear} and the evidence sentences retrieved by ESIM, among which the first two are required evidence pieces.
Figure~\ref{fig:case:token} presents the distribution of attentions from the first evidence to the tokens in the second evidence ($\alpha_i^{2 \rightarrow 1}$) in KGAT (Edge Kernel) and GAT (dot-product). 


The first evidence verifies that ``Al Jardine is an American musician'' but does not enough information about whether ``Al Jardine is a rhythm guitarist''. The edge kernels from KGAT accurately pick up the additional information evidence (1) required from evidence (2): ``rhythm guitarist''. It effectively fills the missing information and completes the reasoning chain. Interesting, ``Al Jardine'' also receives more attention, which helps to verify if the information in the second evidence is about the correct person. This kernel attention pattern is more intuitive and effective than the dot-product attention in GAT. The later one scatters almost uniformly across all tokens and hard to explain how the joint reasoning is conducted. This seems to be a common challenge of the dot-product attention in Transformers~\cite{clark2019does}.

\begin{table}[t]
\centering
\small
\resizebox{0.49\textwidth}{!}{
\begin{tabular}{p{\columnwidth}}
\hline
\textbf{\textcolor{blue}{Claim:}} \textbf{\emph{\textcolor{red}{Al Jardine}}} is an \textbf{\emph{\textcolor{red}{American rhythm guitarist}}}.\\ \hline
\specialrule{0em}{1.5pt}{1.5pt}(1) \textbf{\textcolor{blue}{[Al Jardine]}} \textbf{\emph{\textcolor{red}{Alan Charles Jardine}}} (born September 3, 1942) is \textbf{\emph{\textcolor{red}{an American musician}}}, singer and songwriter who co-founded the Beach Boys.  \\ 
\specialrule{0em}{1.5pt}{1.5pt}(2) \textbf{\textcolor{blue}{[Al Jardine]}} \textbf{\emph{\textcolor{red}{He is best known as the band's rhythm guitarist}}}, and for occasionally singing lead vocals on singles such as ``Help Me, Rhonda'' (1965), ``Then I Kissed Her'' (1965) and ``Come Go with Me'' (1978).  \\
\specialrule{0em}{1.5pt}{1.5pt}(3) \textbf{\textcolor{blue}{[Al Jardine]}} In 2010, Jardine released his debut solo studio album, A Postcard from California.  \\
\specialrule{0em}{1.5pt}{1.5pt}(4) \textbf{\textcolor{blue}{[Al Jardine]}} In 1988, Jardine was inducted into the Rock and Roll Hall of Fame as a member of the Beach Boys.  \\
\specialrule{0em}{1.5pt}{1.5pt}(5) \textbf{\textcolor{blue}{[Jardine]}} Ray Jardine American rock climber, lightweight backpacker, inventor, author and global adventurer. \\
\hline
\textbf{\textcolor{blue}{Label:}} SUPPORT \\
\hline
\end{tabular}}
\caption{\label{tab:case}An example claim~\cite{zhou2019gear} whose verification requires multiple pieces of evidence.}
\end{table}

\section{Conclusion}
This paper presents KGAT, which uses kernels in Graph Neural Networks to conduct more accurate evidence selection and fine-grained joint reasoning.
Our experiments show that kernels lead to the more accurate fact verification.
Our studies illustrate the two kernels play different roles and contribute to different aspects crucial for fact verification.
While the dot-product attentions are rather scattered and hard to explain, the kernel-based attentions show intuitive and effective attention patterns: the node kernels focus more on the correct evidence pieces; the edge kernels accurately gather the necessary information from one node to the other to complete the reasoning chain.
In the future, we will further study this properties of kernel-based attentions in neural networks, both in the effectiveness front and also the explainability front.


\section*{Acknowledgments}
This research is jointly supported by the NSFC project under the grant no. 61661146007, the funds of Beijing Advanced Innovation Center for Language Resources (No. TYZ19005), and the NExT++ project, the National Research Foundation, Prime Minister’s Office, Singapore under its IRC@Singapore Funding Initiative.
\normalsize
\balance
\bibliography{citation}
\bibliographystyle{acl_natbib}
\end{document}